\documentclass[letterpaper, 10 pt, conference]{ieeeconf}
\IEEEoverridecommandlockouts
\usepackage{cite}
\usepackage{amsmath,amssymb,amsfonts}
\usepackage{algorithmic}
\usepackage{graphicx}
\usepackage{textcomp}
\usepackage{xcolor}
\usepackage{fancyhdr,graphicx,amsmath,amssymb}
\usepackage[ruled,vlined]{algorithm2e}
\include{pythonlisting}
\usepackage{balance}
\usepackage[T1]{fontenc}

\def\BibTeX{{\rm B\kern-.05em{\sc i\kern-.025em b}\kern-.08em
		T\kern-.1667em\lower.7ex\hbox{E}\kern-.125emX}}
		
\DeclareMathOperator{\atantwo}{atan2}

\begin{document}
	
\pagestyle{empty}  
\thispagestyle{empty} 

\title{\LARGE \bf
Fast Adaptable Mobile Robot Navigation in Dynamic Environment\\
\thanks{Authors are with Department of Electrical and Computer Engineering and Computer Science, University of Detroit Mercy, Detroit, MI, 48221 U.S., e-mail: (maxi3, sunho3, xuen, cuiso, yinbo, faiedma)@udmercy.edu.}
}

\author{Xihan Ma, Honglin Sun, Enwei Xu, Song Cui, Boqun Yin, Mariam Faied}

\maketitle




\begin{abstract}
	Autonomous navigation in dynamic environment heavily depends on the environment and its topology. Prior knowledge of the environment is not usually accurate as the environment keeps evolving in time. Since robot is continuously evaluating the environment as it proceeds, deciding the optimal way to traverse the environment to get to the goal, computationally efficient yet mathematically adaptive navigation algorithms are needed. In this paper, a navigation scheme for mobile robot, capable of dealing with time variant environment is proposed. This approach consists of a global planner (A*) and local planner (VFH) to assure an optimal and collision-free robot motion. The algorithm is tested both in simulation and experimentation in different environments that are known to result in failures in VFH and ROS navigation stack, for comparison purposes. Overall, the algorithm enables the robot to get to the goal faster and also produces a smoother path while doing so.
\end{abstract}


\section{Introduction}
Due to a booming number of robotic applications that necessitates Autonomous Mobility Robots (AMR), an effective navigation approach that assures optimal robot motion to the desired position while avoiding obstacles is significantly important. Since in reality, the environment is often time-variant, the robot should also be able to handle dynamic obstacles and maximally stick to the shortest path. This paper is going to mainly focus on developing a motion planning scheme that allows the robot to travel in dynamic environment.

\subsection{Related Work}
Navigation in dynamic environment is a popular topic in recent years. the state-of-art approaches adopt reinforcement learning to obtain the optimal policy that deals with moving obstacles\cite{b1}. Mohanan et al.\cite{b2} divided commonly-seen navigation approaches into artificial potential field based, velocity based, probabilistic based etc., however, this classification can be further simplified to local based, global based, global $\&$ local hybrid based algorithms. 

Global navigation requires a map to run graphic search algorithms that output a collision-free path from the starter to the goal. In this case, A* and D* are two frequently used path planner. An algorithm that combines Depth-first Search (DFS) and Breadth-first Search (BFS) to achieve fast and optimal global path plan is proposed in \cite{b3}. In order to suit for time-evolving map, S. Koenig et al. proposed a incremental version of D* algorithm that automatically re-calculates global path once detecting environmental change\cite{b4}. Later the equivalent version of A* was proposed in \cite{b5} that continuously searches for the shortest path to handle pop-up obstacles. Nevertheless, digging into sophisticated real-time graphic search methods leads to unnecessary computational burden and may cause jerky movement if the re-plan is triggered too frequently, a simple but effective algorithm is still preferred.

A well-known local based navigation is Vector Field Histogram (VFH) approach proposed by J. Borenstein and Y. Koren \cite{b6}. This algorithm is a refined version of potential field approach. It acquires sensory data to create obstacle polar histogram, from which the robot picks the optimal admissible valley to steer. Enhanced VFH was proposed by K Balan et al\cite{b7} where VFH is coupled with accessibility graph through Fuzzy Logic to obtain smoother motion and prevent local-minima. Other approaches include probabilistic mapping on the dynamics of the obstacle and calculates collision-free velocity for the robot\cite{b8}. Yet, planning entirely under local perspective may lose global optimality and often performs poorly when encountering situation like U-trap, narrow corridor, etc.

Based on the factors above, a local and global hybrid navigation strategy is more effective. The method adopted in ROS (Robot Operating System) is a cost-map based navigation that takes advantage of a global planner (Dijkstra) and a local planner (DWA) to achieve real-time updated navigation. The former is an efficient graphical search algorithm similar to A*, while the latter takes samples of all possible velocity command pair in a dynamic window and select the one that minimizes the cost function via forward propagation. However, this navigation scheme is sophisticated when deploying. Moreover, conflicts between the two planners could arise under certain circumstances.
\subsection{Contribution}
This paper proposed a light-weight but robust hybrid robot navigation algorithm that combines VFH and A* suitable for time variant environment. The incorporation of VFH and A* for mobile robot navigation is novel in the field. Both simulation and hardware experiments show that our approach leads to successful and smooth mobile robot navigation. The global and local planner compensates for each other and there's little chance for the planners to conflict with each other. The rest of the paper is organized as follows : Section II introduces the mathematical model of the robot kinematics and environment representation, Section III details the proposed algorithm, Section IV demonstrates and explains the results we obtained from both simulation and hardware test under various scenarios and Section V draws the conclusion accordingly.
\section{Modeling}
We developed mathematical modeling for our robot and its surrounding environment:

\subsection{Environment Modeling}
\begin{figure}[ht!]
    \centering
    \includegraphics[scale=0.6]{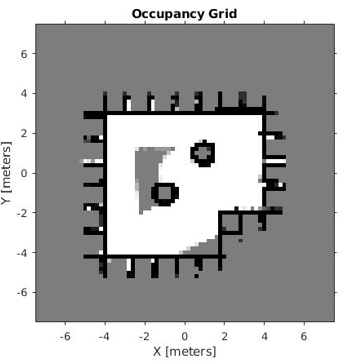}
    \caption{Occupancy Grid Mapping of the Testing Environment: pixels in while are free areas, pixels in black are obstacles and pixels in gray are unexplored}
    \label{fig:map1}
\end{figure}

We apply probabilistic occupancy grid map implemented in \cite{b9} to model the environment. This approach decouples the environment of given size (estimation of the actual field size) into multiple cells and assigns unique values to cells representing different state, namely, 0 for "free", 1 for "occupied" and -1 for "unexplored" (shown in Fig.\ref{fig:map1} as an example). This map gets updated using laser scanner data. Introducing probabilistic methodology into mapping compensates for the odometry error and LiDAR scanning error resulting from wheel slippery and uneven ground, both of which cause accumulating offsets in the map.

\subsection{Robot Modeling}
\begin{figure}[ht!]
    \centering
    \includegraphics[scale = 0.18]{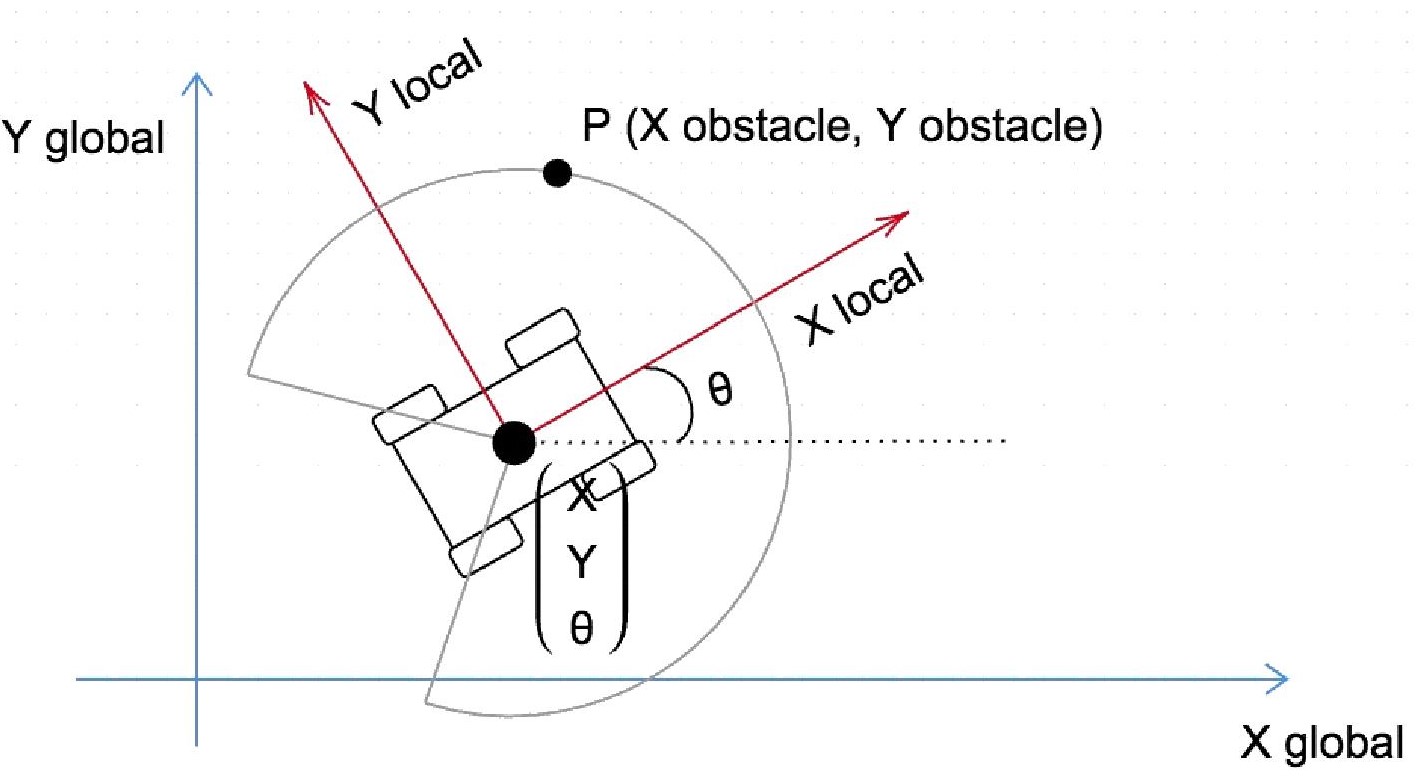}
    \caption{Wheeled Robot Coordinate Frame Convention}
    \label{fig:robot_model}
\end{figure}

Consider a planar mobile robot shown in Fig.\ref{fig:robot_model}, a range scanner is mounted on top of the robot to detect surrounding obstacles and return the distance to the obstacle with respect to the robot's local frame. The robot states representing its position and orientation expressed under the global frame is:
\begin{equation}
    \begin{bmatrix} x,  y,  \theta \end{bmatrix}^\intercal
\end{equation}
To achieve trajectory tracking, given the instantaneous reference state $[x_{ref}, y_{ref}, \theta_{ref}]^\intercal$, we adopted the error dynamics introduced in \cite{b10}:
\begin{equation}
    \begin{bmatrix} {e_1} \\ {e_2} \\ {e_3} \end{bmatrix}
    =
    \begin{bmatrix} \cos\theta & \sin\theta & 0 \\ -\sin\theta & \cos\theta & 0 \\ 0 & 0 & 1 \end{bmatrix}
    \cdot
    \begin{bmatrix} x_{ref}-x \\ y_{ref}-y \\ \theta_{ref}-\theta \end{bmatrix}
\end{equation}
Here, $e_1$, $e_2$ and $e_3$ are the tracking errors to be eliminated. The controls of the robot are its linear velocity $v$ and the angular velocity $\omega$ which can be obtained through:
\begin{equation}
    v = v_d \cos e_3 - u_1
\end{equation}
\begin{equation}
    w = w_d-u_2
\end{equation}
where $v_d$ and $w_d$ are the instantaneous desired linear and angular velocities, calculated as:
\begin{equation}
    v_d = \sqrt{\dot x_{ref}^2 + \dot y_{ref}^2}
    \label{vd}
\end{equation}
\begin{equation}
    w_d = \frac{\ddot y_{ref} \dot x_{ref}-\ddot x_{ref} \dot y_{ref}}{\dot x_{ref}^2 + \dot y_{ref}^2}
    \label{wd}
\end{equation}
We adopt the non-linear control law verified and implemented in \cite{b10}:
\begin{equation}
    u_1 = -k_1 e_1
    \label{u1}
\end{equation}
\begin{equation}
    u_2 = -k_2 v_d \frac{\sin e_3}{e_3} - k_3 e_3
    \label{u2}
\end{equation}
where $k_1$ $k_2$ and $k_3$ are positive definite gains. Lastly, our robot is differential-drive, therefore the instantaneous radius of turning $R$ is determined by:
\begin{equation}
    R = \frac{v}{\omega}
    \label{diff_drive}
\end{equation}

\section{Proposed Algorithm}
\begin{figure}[ht!]
    \centering
    \includegraphics[scale=0.5]{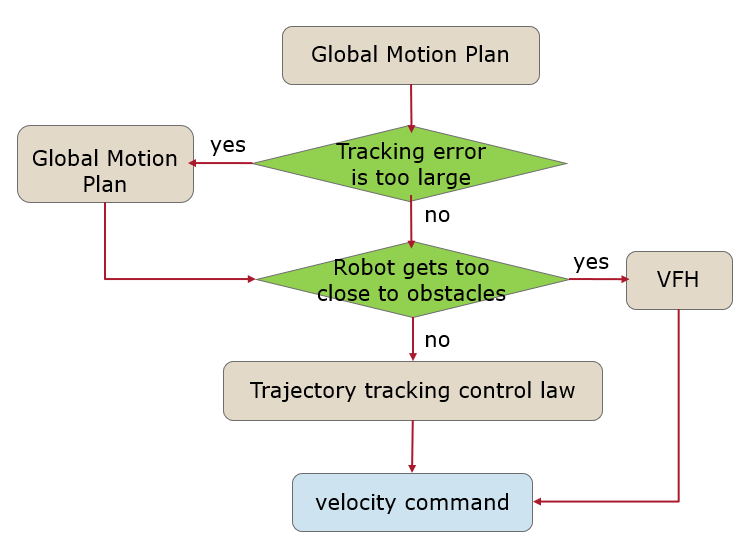}
    \caption{Dynamical Obstacle Avoidance \& Navigation Flow Chart}
    \label{fig:flowchart}
\end{figure}

The flowchart of the proposed dynamical obstacle avoidance algorithm is shown in Fig. \ref{fig:flowchart}.
A global motion plan is generated at the beginning of the navigation, providing a reference trajectory for the robot. If the tracking error is greater than a small enough threshold, motion planning will be re-executed. The robot will mostly governed by the non-linear control law stated in section II but keep reading the distance to the nearest obstacle from the LiDAR and switch to VFH speed command if the obstacle is too close.

The global motion plan sits on the prerequisite that the robot already possessed map of the environment and localized itself at $t=0$. After the desired position and orientation has been specified by the user, the initial conditions (map, start and goal) are fed into the path planner where A* generates a collision-free path. Then the path is sent into the trajectory planner to obtain a function of time quintic trajectory. The reference velocity and acceleration are calculated by finding the first order and second order derivative of the trajectory accordingly. With the reference position, velocity and acceleration at each time sample, the instantaneous linear and angular velocity can be computed using Eqs. (\ref{u1}) - (\ref{u2}).

\subsection{Path Planning}
Based on the environment modeling, we use A* path planner to search for a collision free path. A* takes the current pose, the goal pose, as well as the occupancy grid as inputs, computes an optimal path that does not pass through any obstacles. 
The resultant path is $n$ via-points represented by two vectors $X_{n\times1}$ and $Y_{n\times1}$:
\begin{equation}
    X = [x_1 \quad x_2 \quad x_3 \quad ... \quad x_n]
\end{equation}
\begin{equation}
    Y = [y_1 \quad y_2 \quad y_3 \quad ... \quad y_n]
\end{equation}

\subsection{Trajectory Planning}
The objective for trajectory planning is to find a polynomial time function to fit A* path. To find a balance between reasonable computation amount and a close fit to the path plan, we decided to use quintic polynomial:
\begin{equation}
    x_{ref} = a_{x0} + a_{x1} t + a_{x2} t^2 + a_{x3} t^3 + a_{x4} t^4 + a_{x5} t^5
\end{equation}
\begin{equation}
    y_{ref} = a_{y0} + a_{y1} t + a_{y2} t^2 + a_{y3} t^3 + a_{y4} t^4 + a_{y5} t^5
\end{equation}
\begin{equation}
    \theta_{ref} = \atantwo(\dot y_{ref}, \dot x_{ref}) + k\pi \quad k = 0, 1
\end{equation}
Denote 
\begin{equation}
    a_x = [a_{x0} \quad a_{x1} \quad a_{x2} \quad a_{x3} \quad a_{x4} \quad a_{x5}]^\intercal
\end{equation}
\begin{equation}
    a_y = [a_{y0} \quad a_{y1} \quad a_{y2} \quad a_{y3} \quad a_{y4} \quad a_{y5}]^\intercal
\end{equation}
the goal now is to find $a_x$ and $a_y$. Firstly, we specify the end-time by which the robot should arrive at the goal according to how far is the robot from the goal:
\begin{equation}
    t_f = T \cdot Range
\end{equation}
where $Range$ is the Cartesian distance from the robot's current position to the goal position, $T$ is a positive time scaling factor. Given $n$ via-points assuming the robot travels along the trajectory at a constant speed, the time samples are:
\begin{equation}
    time = [t_0 \quad t_1=t_0+\delta t \quad t_2=t_1+\delta t \quad ... \quad t_f]
\end{equation}
where 
\begin{equation}
    \delta t = \frac{t_f}{n}
\end{equation}
Plugging time samples into the quintic function gives a matrix $A$:
\begin{equation}
    A = \begin{bmatrix} 1 \quad t_0 \quad t_0^2 \quad t_0^3 \quad t_0^4 \quad t_0^5 \\ 0 \quad 1 \quad 2 t_0 \quad 3 t_0^2 \quad 4 t_0^3 \quad 5 t_0^4 \\ ... \\ 1 \quad t_f \quad t_f^2 \quad t_f^3 \quad t_f^4 \quad t_f^5 \\ 0 \quad 1 \quad 2 t_f \quad 3 t_f^2 \quad 4 t_f^3 \quad 5 t_f^4\end{bmatrix}
\end{equation}
we can further define two vectors $b_x$ and $b_y$ that specify the via-points on the trajectory, as well as the desired horizontal and vertical velocities at these via-points:
\begin{equation}
    b_x = [x_1 \quad v_{x_1} \quad x_2 \quad v_{x_2} \quad ... \quad x_n \quad v_{x_n}]^\intercal
    \label{bx}
\end{equation}
\begin{equation}
    b_y = [y_1 \quad v_{y_1} \quad y_2 \quad v_{y_2} \quad ... \quad y_n \quad v_{y_n}]^\intercal
    \label{by}
\end{equation}
Therefore, $a_x$ and $a_y$ can be calculated by:
\begin{equation}
    a_x = A^\dagger b_x
\end{equation}
\begin{equation}
    a_y = A^\dagger b_y
\end{equation}
and the quintic trajectory is obtained.

\subsection{Local Obstacle Avoidance}

As  shown in Fig. \ref{fig:flowchart}, VFH is triggered when the robot is too close to the obstacle. Yet, in order to maximally stay on the desired trajectory, at any time instance $t$ and robot configuration $x$, $y$, $\theta$ at $t$, the target position for VFH under global frame should be $x_{ref}(t)$, $y_{ref}(t)$. Converting this position to robot frame gives current goal position $x_{local}$, $y_{local}$ with respect to local frame:
\begin{equation}
    \begin{bmatrix} x_{local} \\ y_{local} \end{bmatrix} = \begin{bmatrix} \cos\theta \quad \sin\theta \\ -\sin\theta \quad \cos\theta \end{bmatrix} \cdot (\begin{bmatrix} x_{ref}(t) \\ y_{ref}(t)\end{bmatrix} - \begin{bmatrix} x \\ y \end{bmatrix})
\end{equation}
then the target direction $\theta_{tar}$ for VFH is:
\begin{equation}
    \theta_{tar} = \atantwo(y_{local},x_{local})
\end{equation}

The robot's velocity need to be determined according to the steering direction. For the purpose of smoothing the robot's movement, we set its linear velocity as a constant (0.2 [m/s] for our case), and only change its angular velocity.
\begin{figure}[!ht]
	\vspace{-6 pt}
	\centering
	\includegraphics[width=0.7\linewidth]{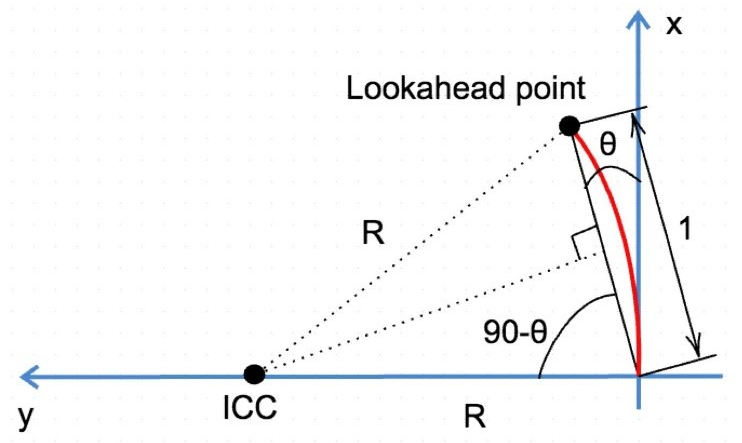}
    	\caption{Angular Velocity Calculation}
	\label{fig:Angular-velo}
\end{figure}
In determining the angular velocity, consider a look-ahead point along the steering direction, 1-unit distance away from the origin of the robot local frame, as is shown in the figure above, where {$\theta$} is the steering direction. We intend to find an arc that links the local frame origin to the look-ahead point as the robot’s desired trajectory. It is easily known from the graph that:
\begin{equation}
    R\cos(90-\theta)=\frac{1}{2}
\end{equation}
where $R$ is the arc radius. Therefore: 
\begin{equation}
    R=\frac{1}{2\sin(\theta)}
    \label{purepursuit}
\end{equation}

Again, we assume the robot to be differential steered, therefore by integrating Equ. (\ref{diff_drive}) into Equ. (\ref{purepursuit}) we obtain:
\begin{equation}
    \omega=2v\sin(\theta)
\end{equation}

During implementing the VFH algorithm, we notice drastic oscillation occurs. To smooth the movement of the robot, we apply a first order IIR filter to reduce the high frequency change in robot’s angular velocity. The difference equation of the filter is given by:
\begin{equation}
    \omega[n]=a_1 \omega_{prev}[n-1]+b_0 \omega_{new}[n]
\end{equation}
where {$\omega_{prev}[n-1]$} is the last angular velocity, {$\omega_{new}[n]$} is the new angular velocity command, {$\omega[n]$} is the new angular velocity to be sent to the robot.

\section{Results}

\par
\begin{figure}[ht!]
    \centering
    \includegraphics[scale=0.5]{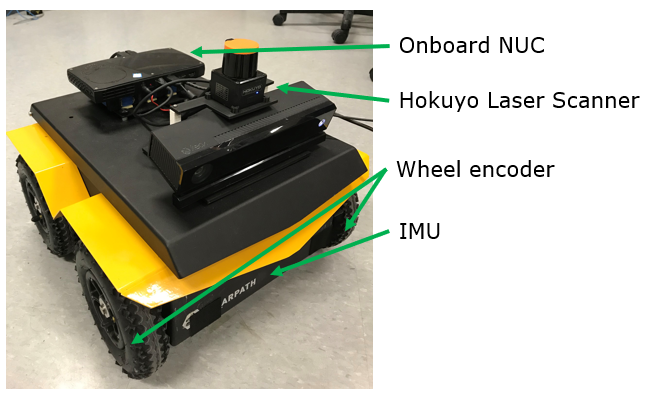}
    \caption{Jackal UGV hardware setup}
    \label{fig:jackal}
\end{figure}

 Simulation and hardware testing are performed to validate the proposed algorithm and compare with counterpart navigation scheme from ROS navigation package. The hardware used for experiment is a Jackal UGV mounted with Hokuyo LiDAR and on board NUC (see Fig.\ref{fig:jackal}). The robot localizes itself using the EKF fused data from the wheel encoder and the embedded IMU sensor. The Hokuyo LiDAR on top of the robot provides 270 degree scanning range and returns a 1080 by 1 matrix indicating distance to the obstacles at every 0.25 degree with respect to local axis.

\subsection{Algorithm Verification}
\begin{figure}[ht!]
    \centering
    \includegraphics[width=8.5cm]{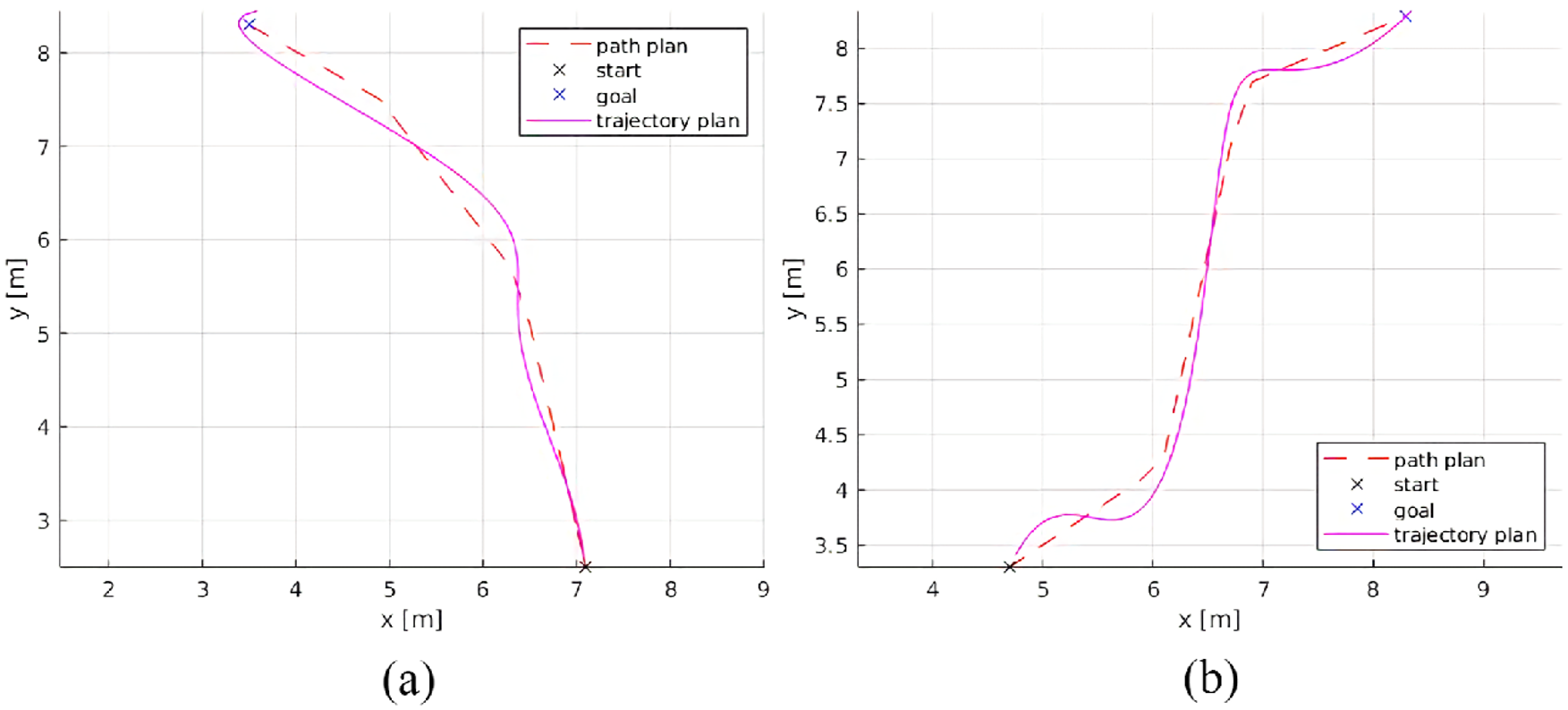}
    \caption{Trajectory Plan Examples ((a) and (b) show two difference environment, obstacles are not displayed here)}
    \label{fig:traj_plan}
\end{figure}

We tested the trajectory tracking control law in a customized Gazebo world simulator. Taking the A* path as input, the trajectory plan is demonstrated in Fig. \ref{fig:traj_plan} where the trajectory is shown on top of the A* path for comparison purpose. The desired linear and angular velocity can be derived from trajectory using Eqs. (\ref{vd}) - (\ref{wd}).

\begin{figure}[ht!]
    \centering
    \includegraphics[width=8.5cm]{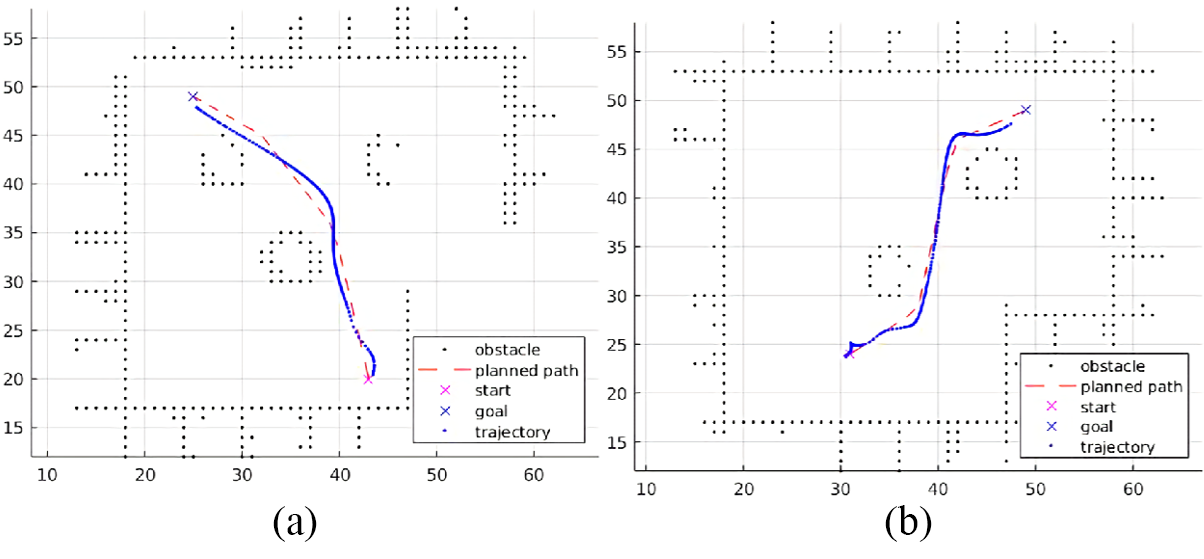}
    \caption{Trajectory Execution Examples: obstacles are black dots, the blue curve shows the recorded robot position, the red curve shows the A* path}
    \label{fig:traj_exe}
\end{figure}

Fig. \ref{fig:traj_exe} shows the robot executing the planned trajectory in a static environment. Here, the obstacles are marked with black dots. One can tell that with the global motion planner alone, the robot is able to safely navigate itself to the goal. 

\begin{figure}[ht!]
    \centering
    \includegraphics[width=8.5cm]{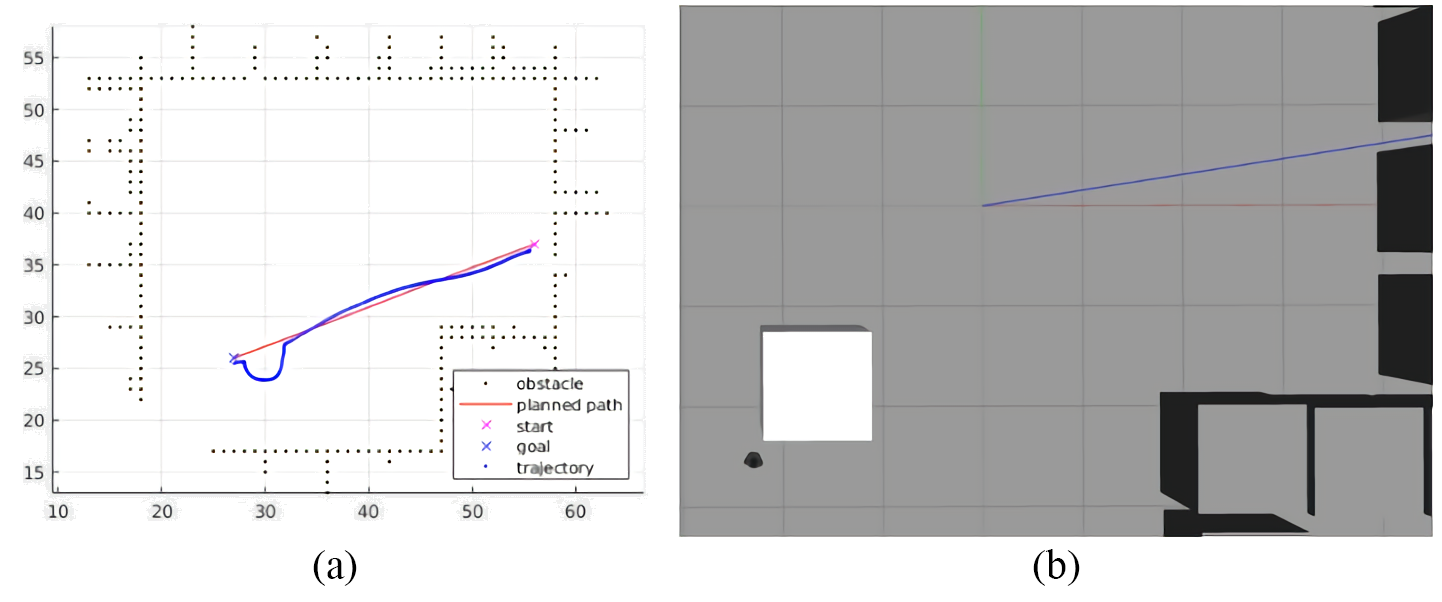}
    \caption{Avoiding Pop-up Obstacle with VFH: (a) shows the robot recorded position and the A* path; The while box in (b) is the pop-up obstacle deployed after mapping the environment}
    \label{fig:VFH}
\end{figure}

Fig.\ref{fig:VFH} shows an example of VFH kicks in when the robot detects the obstacle that didn't appear during map building.

\begin{figure}[ht!]
    \centering
    \includegraphics[width=8.5cm]{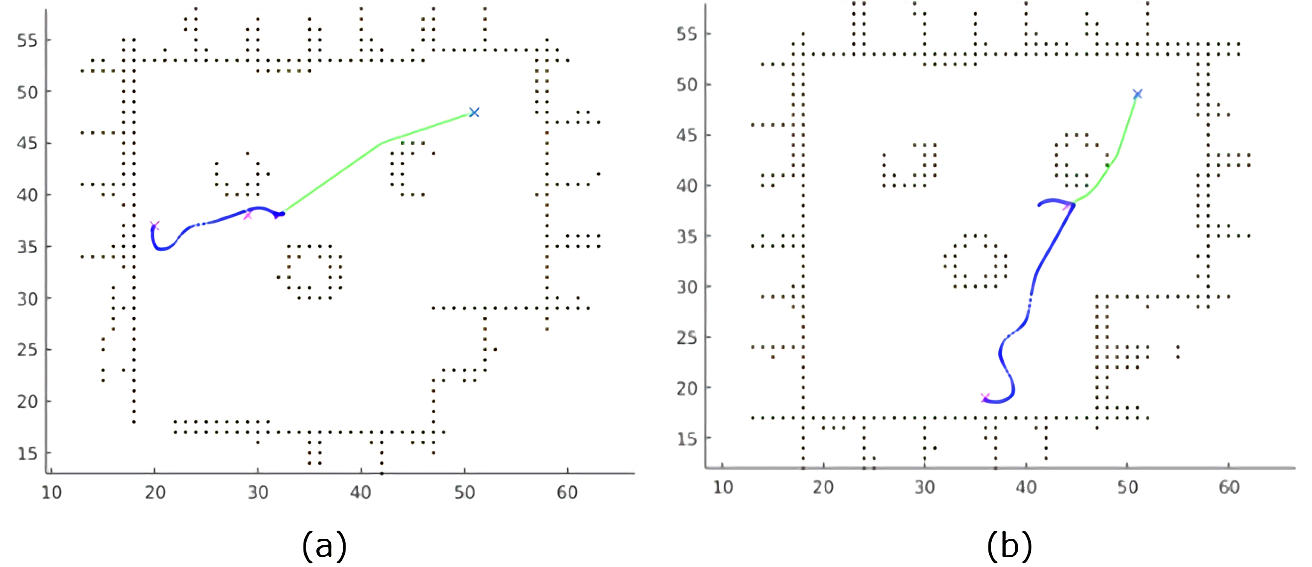}
    \caption{Motion Replan Triggered Examples (the blue curve is the recorded robot position, the green curve is the new path plan based on the latest environment info)}
    \label{fig:replan}
\end{figure}

Nevertheless, after switching to VFH, the tracking errors will start growing. When the errors get too large and the robot is still not able to get rid of the obstacle, the motion planner will be triggered again. (see Fig.\ref{fig:replan})

\begin{figure}[ht!]
    \centering
    \includegraphics[width=8cm]{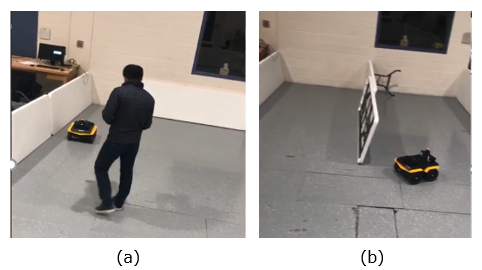}
    \caption{Hardware Experiment Setup: (a) shows the operator moves to block the robot, (b) shows the robot navigating through a static environment}
    \label{fig:hardware_setup}
\end{figure}

We tested our algorithm on the Jackal UGV platform both in static and dynamic environment (see Fig.\ref{fig:hardware_setup}). The static environment consisted of walls in a U-shape. They robot needs to travel from one dead end to the other, while in the dynamic environment, the tester kept walking in the field to block the robot which was traveling from one corner to another. 

\begin{figure}[ht!]
    \centering
    \includegraphics[width=8cm]{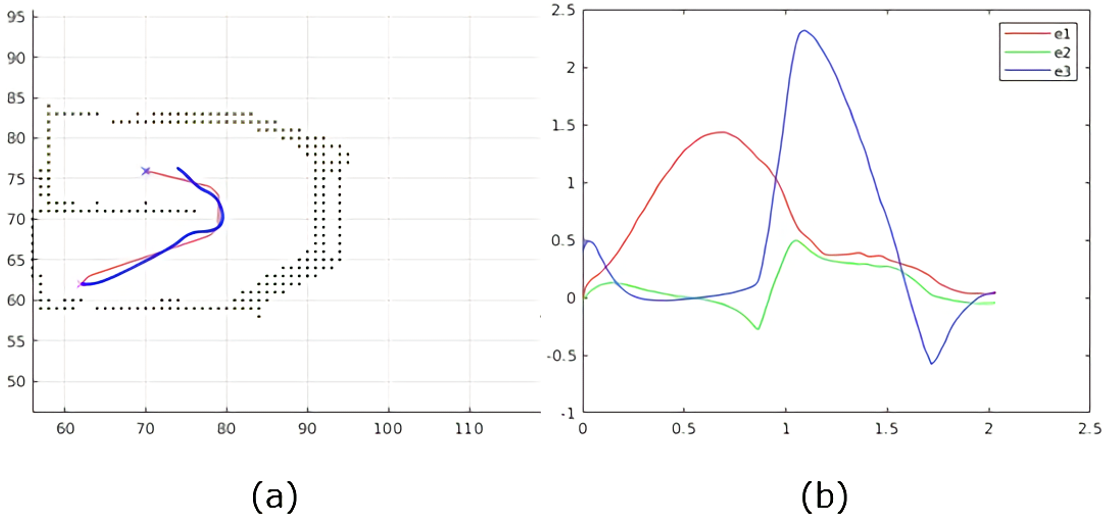}
    \caption{Hardware Experiment in Static Environment: (a) shows the recorded robot position inside the map, (b) shows the errors vs. time}
    \label{fig:hw1}
\end{figure}

The results obtained in static environment is shown in Fig.\ref{fig:hw1}. We can tell from the robot's trajectory that it mostly followed the A* path plan without having VFH kicked in and no re-plan was triggered as well. The errors also dropped to a trivial level as the robot reached the goal.

\begin{figure}[ht!]
    \centering
    \includegraphics[width=8cm]{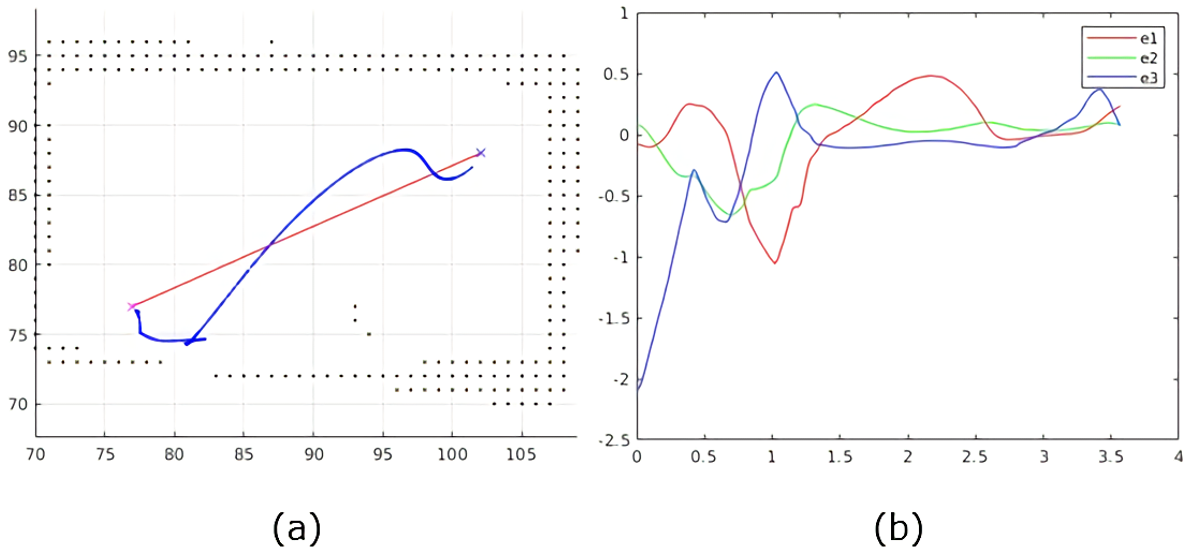}
    \caption{Hardware Experiment in Dynamic Environment: (a) shows the recorded robot position inside the map, (b) shows the errors vs. time}
    \label{fig:hw2}
\end{figure}

For dynamic environment testing, the robot first executed global motion plan without the operator's presence. When the robot started moving, the operator entered the field and tried repeatedly to block the robot (see Fig.\ref{fig:hw2}). From the results, we can tell re-plan was triggered from the discrepency in the robot's trajectory, and there are slight fluctuation in the errors due to the operator's blocking behavior. However, the robot successfully arrived at the goal eventually.

\subsection{Comparing with ROS NAV-STACK}
\begin{figure}[ht!]
    \centering
    \includegraphics[width=8cm]{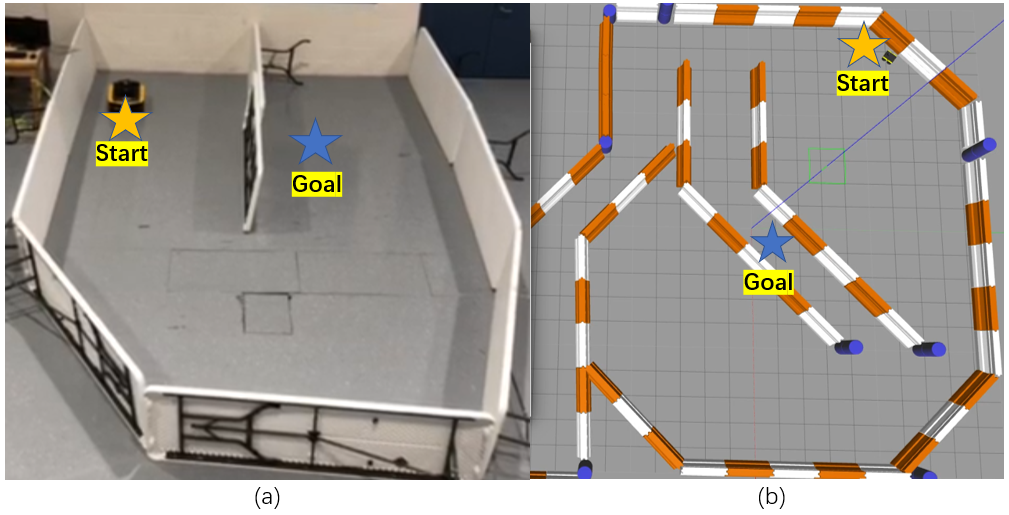}
    \caption{(a) Experiment field setup; (b) simulation field setup}
    \label{fig:field_setup}
\end{figure}

\begin{figure}[ht!]
    \centering
    \includegraphics[width=8cm]{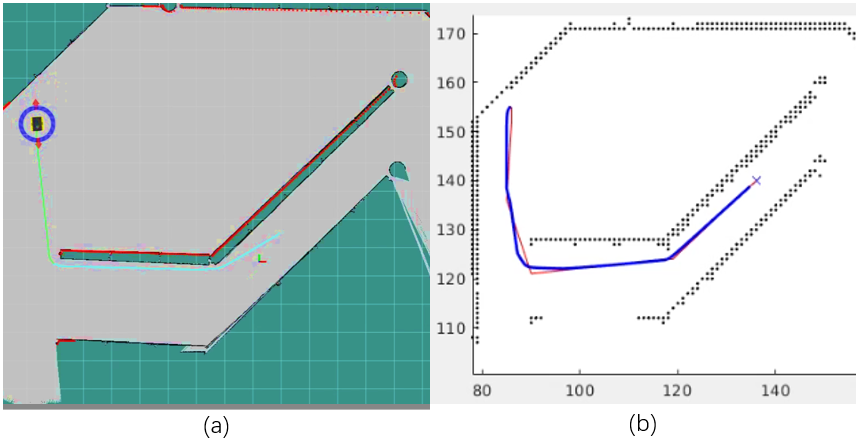}
    \caption{Simulation comparison; (a) ROS NAV-STACK global plan; (b) VFH-A* global plan}
    \label{fig:sim_cmp}
\end{figure}

\begin{figure}[ht!]
    \centering
    \includegraphics[width=8cm]{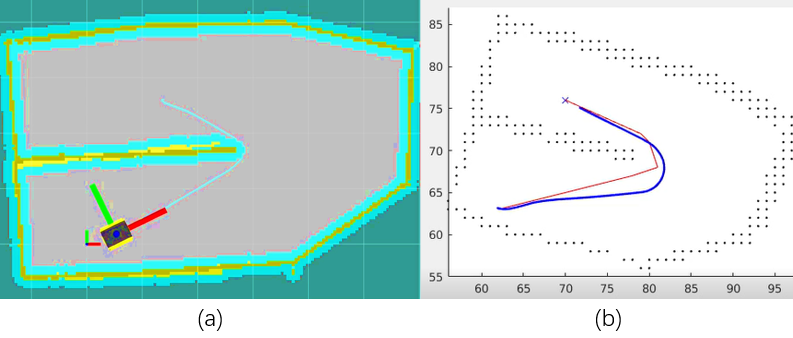}
    \caption{Experiment comparison: (a) ROS NAV-STACK global plan; (b) VFH-A* global plan}
    \label{fig:hw_cmp}
\end{figure}

We compared our proposed navigation scheme with the ROS navigation stack.
The field setup for comparison is seen in Fig.\ref{fig:field_setup}. The results are demonstrated in Fig.\ref{fig:sim_cmp} and Fig.\ref{fig:hw_cmp}. 

The robot's running time is shown in Table \ref{cmp_tab}.

\begin{table}[ht!]
	\vspace{-6 pt}
	\caption{Run-time Comparison}
	\begin{center}
		\begin{tabular}{|c|c|c|c|}
			\hline
			\ & ROS NAV-STACK & VFH-A* \\
			\hline
			Simulation & 43 [sec] & 60.98 [sec] \\ 
			\hline
			Experiment & 48.87 [sec] & 36.31 [sec] \\ 
			\hline
		\end{tabular}
		\label{cmp_tab}
	\end{center}
	\vspace{-10 pt}
\end{table}

Because of wider space and better conditioned map in Gazebo simulation, though the global planners gave similar path, ROS NAV-STACK is able to drive the robot faster thanks to the nature of DWA. Yet when it comes to experiment tests using  ROS NAV-STACK, robot stuck at the hairpin turn for several seconds due to the "local minima" issue discussed above. Our proposed method does not seek to maximally track the global path and is able to cope with poor-conditioned map resulting from inaccurate robot localization using VFH. As a result, VFH-A* drove the robot faster in hardware experiment. It is worth noticing that the VFH's linear velocity used in both simulation and experiment are the same, and is set relatively low for safety, therefore the run-time in simulation has great improving room.

\section{Conclusion and Discussion}	
In this paper, we proposed a hybrid navigation scheme for mobile robot that is able to navigate through dynamic environment. It can be concluded that this approach successfully works under various situations both in simulation and hardware test, and drives the robot on a smooth path. It is light-weight, fast, capable of handling pop-up and moving obstacles while maintaining the robot on the optimal path. Comparing to the widely adopted ROS navigation package, there is less chance in the proposed navigation algorithm where the global and local planner conflict, causing to oscillation in the robot motion.

As future work, the algorithm can be further enhanced by optimizing the trajectory to better fit the A* path (e.g adopting spline interpolation between via-points) and seeking to achieve higher velocity to lessen travel time. Tests are needed under more complicated environments (e.g. consider the effect of uneven ground and map the arena in 3D) to show the algorithm's robustness.


\begin{thebibliography}{00}

	\bibitem{b1} M. Everett, Y. F. Chen and J. P. How, "Motion planning among dynamic, decision-making agents with deep reinforcement learning," \emph{2018 IEEE/RSJ International Conference on Intelligent Robots and Systems (IROS)}, Madrid, 2018, pp. 3052-3059
	
    \bibitem{b2} M. Mohanan and A. Salgoankar, “A survey of robotic motion planning
    in dynamic environments,” \emph{Robotics and Autonomous Systems}, vol.
    100, pp. 171–185, 2018.
    
    \bibitem{b3} P. E. Hart, N. J. Nilsson, and B. Raphael, “A formal basis for the
    heuristic determination of minimum cost paths,” \emph{IEEE Transactions
    on Systems Science and Cybernetics}, vol. 4, no. 2, pp. 100–107, 1968.
		
	\bibitem{b4} S. Koenig and M. Likhachev, “D* lite,” \emph{Aaai/iaai}, vol. 15, 2002.
		
	\bibitem{b5} S. Koenig, M. Likhachev, and D. Furcy, “Lifelong planning A*,”
    \emph{Artificial Intelligence}, vol. 155, no. 1-2, pp. 93–146, 2004.
    
    \bibitem{b6} J. Borenstein, Y. Koren \emph{et al}., “The vector field histogram-fast obstacle
    avoidance for mobile robots,” \emph{IEEE Transactions on Robotics and
    Automation}, vol. 7, no. 3, pp. 278–288, 1991.
    
    \bibitem{b7} K. Balan, M. P. Manuel, M. Faied, M. Krishnan, and M. Santora,
    “A fuzzy based accessibility model for disaster environment,” \emph{2019
    International Conference on Robotics and Automation (ICRA)}, 2019,
    pp. 2304–2310.
    
    
	\bibitem{b8} C. Fulgenzi, A. Spalanzani, and C. Laugier, “Dynamic obstacle avoidance in uncertain environment combining pvos and occupancy grid,” \emph{Proceedings 2007 IEEE International Conference on Robotics and Automation}, 2007, pp. 1610–1616.
		
	\bibitem{b9} E. Kaufman, T. Lee, and Z. Ai, “Autonomous exploration by expected information gain from probabilistic occupancy grid mapping,” \emph{2016 IEEE International Conference on Simulation, Modeling, and Programming for Autonomous Robots (SIMPAR)}, 2016, pp. 246–251.
	
	
	\bibitem{b10} K. L. Besseghieur, R. Tr˛ebi´nski, W. Kaczmarek, and J. Panasiuk, “Trajectory tracking control for a nonholonomic mobile robot under ros,” \emph{Journal of Physics: Conference Series}, vol. 1016, no. 1, 2018, p. 012008.
	\end{thebibliography}

\end{document}